%% file: RR_multirobot_parkour_arXiv.tex
\documentclass{article}

\usepackage{arxiv}

\usepackage{cite}
\usepackage{amsmath,amssymb,amsfonts}
\usepackage{algorithmic}
\usepackage{graphicx}
\usepackage{textcomp}
\usepackage{xcolor}
\def\BibTeX{{\rm B\kern-.05em{\sc i\kern-.025em b}\kern-.08em
    T\kern-.1667em\lower.7ex\hbox{E}\kern-.125emX}}

\usepackage{subcaption} 

\usepackage{mathtools} 

\usepackage{makecell} 
\usepackage{multirow} 
\usepackage{arydshln} 
\usepackage{booktabs} 
\usepackage{array} 
\renewcommand{\arraystretch}{0.95}
\usepackage{pifont} 
\usepackage{siunitx} 
\newcommand{\figref}[1]{Fig.~\ref{#1}}
\newcommand{\tabref}[1]{Table~\ref{#1}}
\newcommand{\secref}[1]{Sec.~\ref{#1}}
\newcommand{\neqref}[1]{Eq.~\eqref{#1}}

\usepackage[hidelinks]{hyperref}
\usepackage{orcidlink}

\title{
Beyond Egocentric Limits: Multi-View Depth-Based Learning for Robust Quadrupedal Locomotion
}

\author{
    Rémy Rahem \orcidlink{0000-0003-4874-9400} \\
        Department of Electrical and Computer Engineering\\
        Universit\'e de Sherbrooke\\
        Sherbrooke, QC J1N 3C6, Canada \\
        \texttt{remy.rahem@usherbrooke.ca} \\
    \And 
    Wael Suleiman \orcidlink{0000-0002-1968-7207} \\
        Department of Electrical and Computer Engineering, \\
        Universit\'e de Sherbrooke\\
        Sherbrooke, QC J1N 3C6, Canada\\
        \texttt{wael.suleiman@usherbrooke.ca} \\
}

\begin{document}

\maketitle

\markboth{Preprint Version.}
{Rahem \MakeLowercase{\textit{et al.}}: Multi-view depth-based learning for quadruped locomotion} 

\begin{abstract}
    Recent progress in legged locomotion has allowed highly dynamic and parkour-like behaviors for robots, similar to their biological counterparts.
    Yet, these methods mostly rely on egocentric (first-person) perception, limiting their performance, especially 
    when the viewpoint of the robot is occluded.
    A promising solution would be to enhance the robot's environmental awareness by using complementary viewpoints, such as multiple actors exchanging perceptual information.
    
    Inspired by this idea, this work proposes a multi-view depth-based locomotion framework that combines egocentric and exocentric observations to provide richer environmental context during agile locomotion.
    Using a teacher-student distillation approach, the student policy learns to fuse proprioception with dual depth streams while remaining robust to real-world sensing imperfections.
    To further improve robustness, we introduce extensive domain randomization, including stochastic remote-camera dropouts and 3D positional perturbations that emulate aerial-ground cooperative sensing.
    
    Simulation results show that multi-viewpoints policies outperform single-viewpoint baseline in gap crossing, step descent, and other dynamic maneuvers, while maintaining stability when the exocentric camera is partially or completely unavailable.
    Additional experiments show that moderate viewpoint misalignment is well tolerated when incorporated during training.
    
    This study demonstrates that heterogeneous visual feedback improves robustness and agility in quadrupedal locomotion.
    Furthermore, to support reproducibility, the implementation accompanying this work is publicly available at \url{https://anonymous.4open.science/r/multiview-parkour-6FB8}.
\end{abstract}


\input{Sections/introduction}
\input{Sections/related_work}
\input{Sections/methodology}
\input{Sections/results}
\input{Sections/conclusion}
\section{Acknowledgement}
    This work was supported in part by the Natural Sciences and Engineering Research Council of Canada (NSERC) and in part by the Fonds de recherche du Qu\'ebec – Nature et technologies (FRQNT). 

\bibliographystyle{IEEEtran}
\bibliography{multirobot_parkour}

\end{document}

%% file: Sections/introduction.tex
\section{Introduction} \label{sec:introduction}

    Legged robots possess an agility unmatched by wheeled or tracked robots, allowing them to cross over obstacles, steps, and gaps where the latter would fail.
    Such agility comes however with a cost: these robots need robust perception and control, especially when dealing with unstructured environments.
    While model-based control strategies can achieve reliable walking and trotting through precise estimation and planning~\cite{Kuindersma2015,Hutter2016}, the models often lack the adaptability required for dynamic behaviors in real-world scenarios.

    Recent advances in reinforcement learning (RL) have led to agile and adaptive legged locomotion policies without the need of explicit modeling.
    Such controllers can even integrate visual or depth perception directly into policy learning, allowing the robot to exhibit parkour-like agility and traversal over complex terrains~\cite{Cheng2024,Hoeller2024}.
    However, these policies rely primarily on a single egocentric camera view, restricting the environmental awareness of the robot and struggling when visual information becomes unreliable or incomplete, such as when their field of view is occluded.

    One solution to improve the environmental understanding of robots is to combine diverse viewpoints, although such perception is often used only for mapping or planning rather than direct control~\cite{Wisth2023, Gawel2018, Wang2020}.
    Indeed, few works have used multi-view fusion within a learned locomotion policy, leaving open questions about how heterogeneous viewpoints can be exploited directly to realize dynamic motion.
    
    As such, the current work tries to answer some of these questions by proposing a visual locomotion framework for quadruped robot that exploits multiple viewpoints.
    Our approach is based on the teacher-student distillation method used in~\cite{Cheng2024}, extending it to a setting where the student policy learns to combine proprioception with onboard (egocentric) and remote (exocentric) visual inputs.
    
    The contributions of this work are threefold:
        (1)~we propose a two-phase reinforcement learning framework, extending~\cite{Cheng2024} to integrate egocentric and exocentric visual observations;
        (2)~we introduce a domain-randomized training process, which incorporates stochastic visual dropout and camera perturbations to emulate real-world visual uncertainty and improve the robustness of the policy; and
        (3)~we show that multi-viewpoints visual policies outperform single-viewpoint baseline~\cite{Cheng2024} in complex parkour-like environments.
    
    The remainder of this work is organized as follow:
    \secref{sec:related_work} contextualizes our approach by reviewing key developments in learning-based locomotion and multi-viewpoints perception.
    \secref{sec:method} presents the reinforcement learning framework, the stochastic dropout mechanism and camera perturbations, as well as the method used to evaluate the RL policy.
    \secref{sec:results} compares the proposed approach to the baseline~\cite{Cheng2024}.
    Finally, \secref{sec:conclusion} summarizes the main findings of this study and outlines directions for future research.

%% file: Sections/related_work.tex
\section{Related Work} \label{sec:related_work}

This section reviews two key research directions supporting our approach. 
We first outline advances in learning-based locomotion and privileged distillation that enable adaptive, vision-informed control in legged robots. 
We then discuss multi-view and collaborative perception, highlighting how complementary viewpoints enhance spatial reasoning and robustness in dynamic environments.
For each research direction, we compare our approach to the presented methods.

    \subsection{Learning-Based Locomotion and Privileged Distillation} \label{subsec:learning_based_locomotion}
        By enabling autonomous acquisition of locomotion skills through experience rather than explicit modeling, data-driven and learning-based methods, such as RL, modified the way legged robots are controlled.
        Foundational contributions such as~\cite{Schulman2017,Peng2017} demonstrated that stable and adaptive locomotion can emerge directly from optimization, while simulation-to-real (sim-to-real) transfer can be enabled through domain randomization~\cite{Tan2018,Lee2020}.
        These works established that policies trained in simulation can generalize to real quadruped robots, and that proprioceptive-only control can achieve robust locomotion in real environments.
        
        Building on these foundations, adaptive policies, such as \textit{Rapid Motor Adaptation}~\cite{Kumar2021}, have been introduced to estimate the environment dynamics in real-time, leading to policies that could adjust to never-seen before terrain.
        These policies were later extended to extreme agility and parkour-style behaviors~\cite{Hoeller2024,Cheng2024,Zhuang2025}.
        More recently, exploration-driven approaches like \textit{Skill Discovery as Exploration}~(SDAX)~\cite{Rho2025} further expanded these methods by allowing robots to learn diverse skills autonomously, such as crawling, climbing, and leaping, without heavy reward engineering.
        
        Around the same time, privileged learning and distillation techniques also emerged to bridge the sensing gap between simulation and deployment on real robots.
        For example, the \textit{DAGGER} algorithm~\cite{Ross2011} formalized iterative imitation learning, while \textit{Regularized Online Adaptation}~(ROA)~\cite{Fu2022} showed how proprioception history can encode environment dynamics.
        Finally, the authors of~\cite{Agarwal2023} demonstrated how, by combining ROA with a teacher-student distillation method, a teacher policy with access to privileged information can supervise a student policy using only realistic sensory input.
        Subsequent works such as \textit{Vision-Based Terrain-Aware Locomotion}~(ViTAL)~\cite{Fahmi2023}, \textit{Parkour with Implicit-Explicit learning}~(PIE)~\cite{Luo2024}, \textit{extreme-parkour}~\cite{Cheng2024}, \textit{MOVE}~\cite{Li2025}, \textit{Perceptive Internal Model}~(PIM)~\cite{Long2025} and \textit{World Model-based Perception}~(WMP)~\cite{Lai2025} integrated perception directly into end-to-end legged locomotion policies, underlining the growing importance of integrating perception to inform control.
        
        Our method builds on these works by retaining the teacher-student distillation used in \cite{Cheng2024}, but extending it to rely on multiple viewpoints, a dimension unexplored in existing parkour-oriented frameworks.
    
    \subsection{Multi-View and Collaborative Perception}
        While relying on vision becomes increasingly important in recent work on locomotion, most methods remain limited to a single egocentric view. 
        Multi-sensor and collaborative perception aim to overcome this limitation by incorporating complementary viewpoints. 
        Systems such as \textit{visual inertial lidar legged navigation system}~(VILENS)~\cite{Wisth2023} combine vision, inertial, and lidar data for reliable odometry in unstructured terrain, while aerial-ground collaborations~\cite{Wang2020,Gawel2018} demonstrate that external observers enhance spatial awareness for exploration and teleoperation.
        
        Egocentric-exocentric learning has also gained traction, illustrating how combining first- and third-person perspectives can enrich motion and environmental understanding~\cite{Grauman2024,Thatipelli2025} of artificial intelligence agents.

        In contrast, our work focuses on embedding multi-view perception directly into the locomotion policy itself, rather than treating external sensing as a mapping or planning module. 
        By training a locomotion policy that combines egocentric and exocentric visual streams, our approach enables adaptive, viewpoint-aware control for discontinuous terrain.
    

%% file: Sections/methodology.tex
\section{Method} \label{sec:method}
        \begin{figure}[htbp!]
            \centering
            \includegraphics[width=0.5\columnwidth]{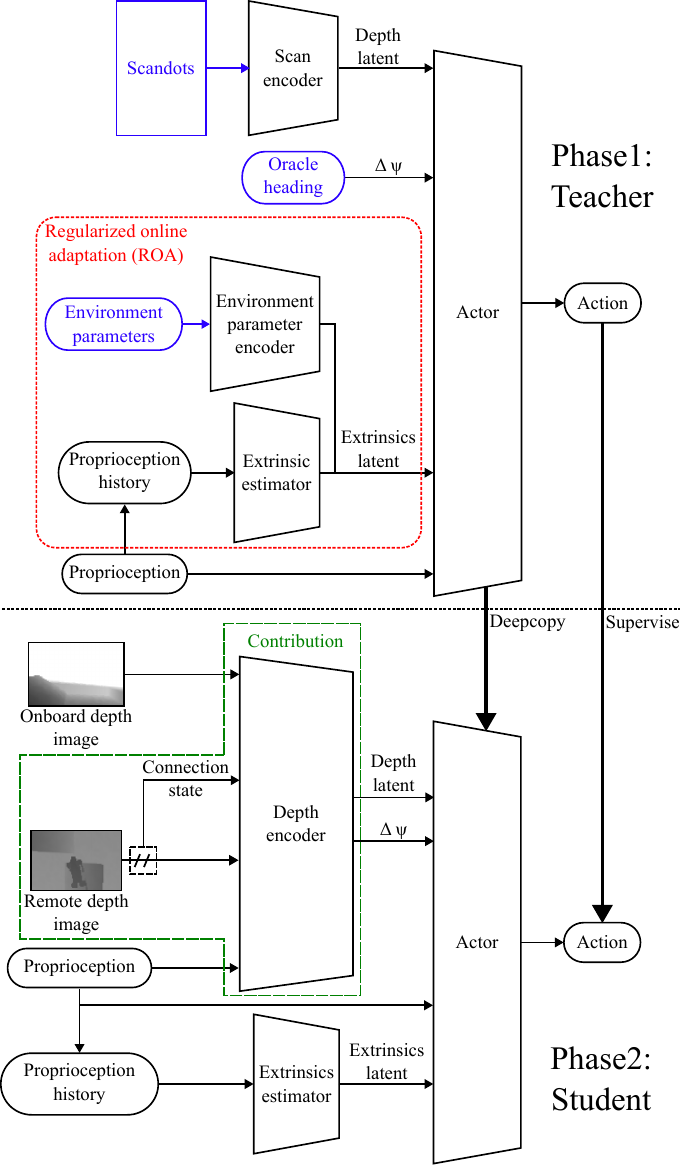}
            \caption{
                The two-phase training: 
                In phase 1, the teacher (top) uses proprioception and privileged information (blue) to train an extrinsic estimator~(EE) and an actor to produce the desired actions, i.e. joint position targets.
                In phase 2, the student (bottom) improves the actor by imitating the teacher using onboard and remote depth streams, proprioception, and the pretrained EE.
            }
            \label{fig:architecture_overview}
        \end{figure}
        
        The proposed approach, illustrated in \figref{fig:architecture_overview}, enables a quadruped robot to traverse discontinuous and visually ambiguous terrain by leveraging multi-view depth perception and a teacher-student distillation framework. 
        Following the approach of \cite{Cheng2024}, a teacher policy is first trained with access to privileged information, such as full environmental states.
        Its behavior is then distilled into a student policy that operates under realistic sensing constraints, relying solely on proprioception and dual depth observations.
        The student thus learns to fuse egocentric and exocentric observations while maintaining robust locomotion under partial perception.
    
    \subsection{Training Environment}\label{subsec:training_environment}
        All experiments are conducted in simulation using the Unitree Go2 quadruped robot within Isaac Gym. 
        The terrain follows the same curriculum as \cite{Cheng2024}, i.e. tilted ramps, gaps, hurdles, and high steps arranged with increasing difficulty.
        
        The physics simulation runs at $f_{sim} = 200\,\mathrm{Hz}$, while the control policy is updated at $f_{ctrl} = 50\,\mathrm{Hz}$, consistent with real hardware. 
        Depth sensing operates at $f_{vis} = 10\,\mathrm{Hz}$, synchronized with every fifth control step.

        \subsubsection{Envrionment pertubations}
        To encourage generalization and sim-to-real robustness, continuous domain randomization is applied on multiple environment values, drawing from multiple independent uniform distributions:
        terrain friction coefficient $\mu\sim\mathcal{U}(0.6,\,2.0)$;
        added mass to the base mass of the robot $\Delta m_{\text{c}}\sim\mathcal{U}(0.0,\,3.0)~\mathrm{kg}$, such that the robot increased mass $m_{\text{c}}'=m_{\text{c}}+\Delta m_{\text{c}}$; and 
        position offset of the robot center of mass (CoM) $\boldsymbol{\Delta} \mathbf{p}_{\text{c}}\sim\mathcal{U}(-0.2,\,0.2)~\mathrm{m}$, such that the robot displaced CoM $\mathbf{p}_{\text{c}}'=\mathbf{p}_{\text{c}}+\boldsymbol{\Delta} \mathbf{p}_{\text{c}}$;
        while actuator strengths are scaled by a factor $\alpha\sim\mathcal{U}(0.8,\,1.2)$. 
        External velocity perturbations $\mathbf{v}_{\text{noise}}\sim\mathcal{U}(0.0,\,0.5)~\mathrm{m/s}$ are applied periodically, i.e. every $8~\mathrm{s}$, to simulate pushes, and a time delay $\delta_t=20~\mathrm{ms}$ is  applied to the visual streams to emulate real-world latency.
    
        \subsubsection{Camera perturbations}
\begin{figure}[htbp!]
                \centering
                \includegraphics[width=0.25\columnwidth]{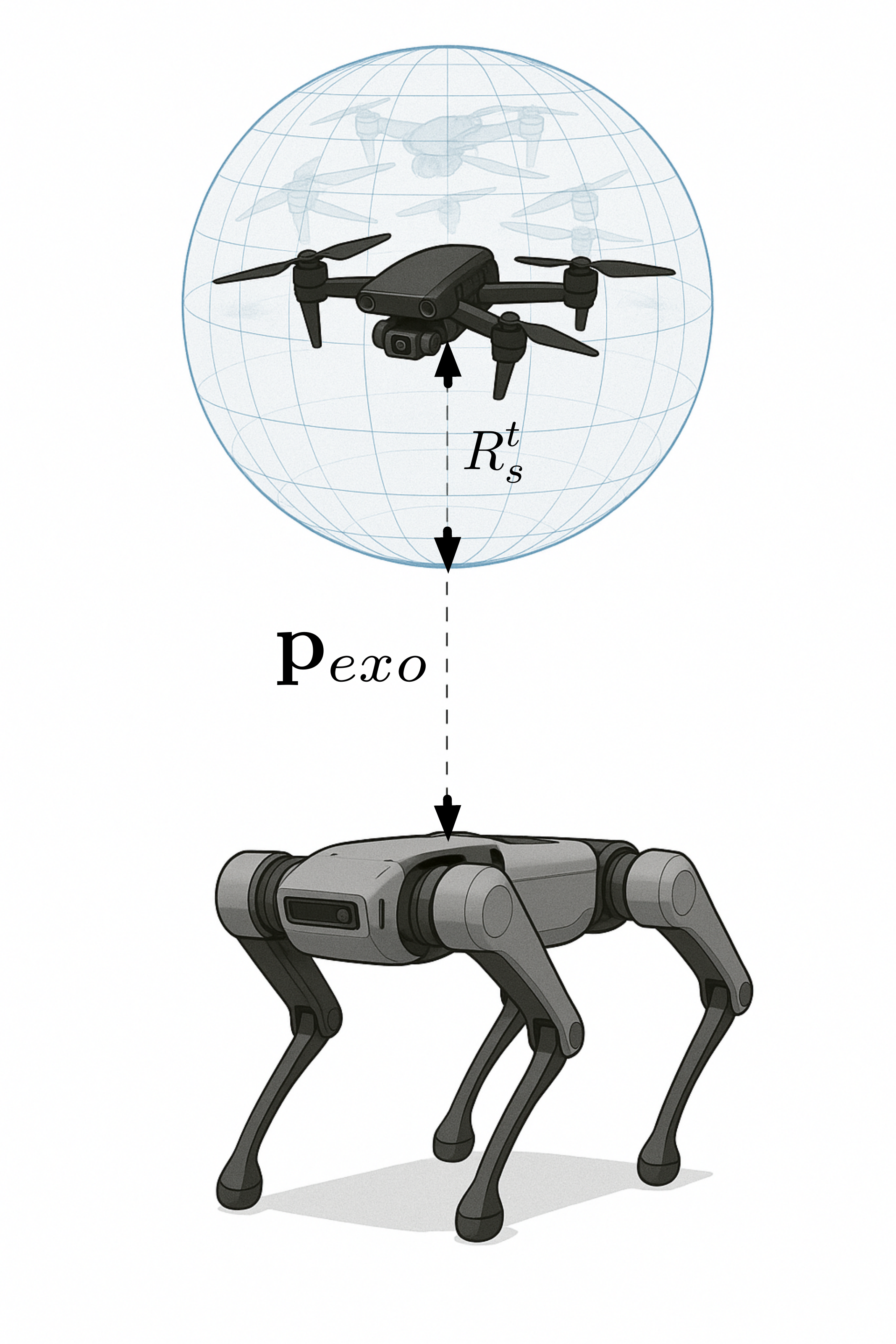}
                \caption{Illustration of the positional randomization of the remote camera within a spherical region of radius $R_s^t$.}
                \label{fig:position_displacement}
            \end{figure}

            The onboard camera pitch angle is randomized as  $\theta_{ego}\sim\mathcal{U}(-5,\,5)^\circ$, 
            whereas the remote camera experiences similar randomization on roll~($\phi_{exo}$), pitch~($\theta_{exo}$), and yaw~($\psi_{exo}$). 
            For positional randomization, as presented in \figref{fig:position_displacement}, we adopt a spherical displacement model~\cite[eq.~(21.5.11)]{Press2007}: 
            \begin{equation}
                \mathbf{p}_{exo}' = \mathbf{p}_{exo} + \mathbf{z}, 
                \qquad
                \mathbf{z} = R_s \sqrt[3]{U} \frac{\mathbf{X}}{\lVert \mathbf{X} \rVert}
            \end{equation}
            where 
            $\mathbf{p}_{exo}', \, \mathbf{p}_{exo}\in\mathbb{R}^3$ are respectively the displaced and nominal position vectors of the remote camera, relative to the quadruped robot CoM; 
            $\mathbf{z}\in\mathbb{R}^3$ is the spherical displacement;
            $\mathbf{X}\in\mathbb{R}^3$ is a vector of independent and identically distributed (i.i.d.) variables for which each element $X_i\sim\mathcal{N}(0,\,1)$; and 
            $U\sim\mathcal{U}(0,\,1)$ is an independent uniformly distributed random variable.
            This ensures uniform sampling within a sphere of radius $R_s$ centered on the nominal pose.
            
            This displacement model simulates a drone hovering imprecisely around the ideal position $\mathbf{p}_{exo}$, carrying the remote camera over the quadruped robot.
            Six models were trained using $R_s \in \{0.0,\,0.1,\,0.2,\,0.3,\,0.4,\,0.5\}~m$ to evaluate the impact of exocentric-view displacement on robustness, with the results presented in \secref{sec:results}.
            
    \subsection{Observations and Input Representation}\label{subsec:observation_and_input_representation}
        At each timestep $t$, the policy receives the following observation vector:
        \[
        \mathbf{o}_t = [\mathbf{o}_t^{\text{prop}}, \, \mathbf{o}_t^{\text{priv}}, \, \mathbf{o}_t^{\text{depth}}, \, b_t]
        \]
        where $\mathbf{o}_t^{\text{prop}}$ is the proprioceptive feedback; 
        $\mathbf{o}_t^{\text{priv}}$ is the privileged information only available to the teacher, but estimated by the student; 
        $\mathbf{o}_t^{\text{depth}}$ is the visual information, coming from either scandots for the teacher policy or onboard and remote depth cameras for the student policy; and 
        $b_t$ is a binary flag indicating the availability of the remote camera.

        \subsubsection{Proprioceptive feedback}
            $\mathbf{o}^{\text{prop}}$ is defined as:
            \[
            \mathbf{o}^{\text{prop}} = [\phi_{\text{c}}, \, \theta_{\text{c}}, \, \mathbf{v}_{\text{c}}, \, \boldsymbol{\omega}_{\text{c}}, \,  \mathbf{q}, \, \dot{\mathbf{q}}, \, \mathbf{C}],
            \]
            where 
            $\phi_{\text{c}}, \, \theta_{\text{c}}\in\mathbb{R}$ are the robot base roll and pitch orientations;
            $\mathbf{v}_{\text{c}}, \, \boldsymbol{\omega}_{\text{c}}\in\mathbb{R}^3$ are the base linear and angular velocities;
            $\mathbf{q}, \, \dot{\mathbf{q}}\in\mathbb{R}^{12}$ are the position and velocity of the joints; and 
            $\mathbf{C}\in\mathbb{R}^4$ are the contact indicators of the feet. 
            Signals are normalized and clipped to ensure numerical stability. 

        \subsubsection{Privileged information}
            $\mathbf{o}^{\text{priv}}$ is a vector made of depth and environmental parameters, to be respectively estimated by student's depth encoder module and Regularized Online Adaptation (ROA)-based Extrinsics estimator (EE) module.
            It is defined as:
            \[
                \mathbf{o}^{\text{priv}} = [\,
                \underbrace{\boldsymbol{\Delta\psi}, \, \Delta h,}_{\mathclap{\substack{\text{Depth} \\\text{parameters}}}} \, \underbrace{\mu, \, m_{\text{c}}', \, \mathbf{p}_{\text{c}}', \, \alpha}_{\mathclap{\substack{\text{Environmental} \\\text{parameters}}}}\
                ]
            \]
            where
            $\boldsymbol{\Delta \psi}\in\mathbb{R}^{2}$ is the oracle heading toward the current goal $\text{g}$ and next goal $\text{g}+1$, such that:  
            \[
                \boldsymbol{\Delta \psi} = [{\Delta \psi_{\text{g}}}, {\Delta \psi_{\text{g+1}}}]
            \]
            and $\Delta h$ is a binary flag indicating if the terrain is flat.
    
        \subsubsection{Visual perception} \label{subsubsec:visual_perception}
            The teacher policy defines $\mathbf{o}^{\text{depth}}$ as scandots located under the robot, i.e. a low resolution elevation map centered on the robot, as proxy for depth images.
            On the other hand, the student policy defines $\mathbf{o}^{\text{depth}}$ as two synchronized depth streams: an onboard egocentric camera and a remote exocentric one. 
            Both depth streams are based on the Intel RealSense D435i depth camera, capturing $106\times60$~px depth images with an $87^\circ$ horizontal field of view, clipped at $3\,\mathrm{m}$.
            Depth values are cropped and resized to \(87\times58\)~px, then the values are normalized and centered around 0 using min-max normalization:
    
            \begin{equation} \label{eq:depth_normalization}
                D' = \frac{D - D_{\min}}{D_{\max} - D_{\min}} -0.5\\
            \end{equation}
    
            Where $D'$, $D$, $D_{\min}$ and $D_{\max}$ represent the normalized depth value, the raw depth value, and the lower and upper limits of the clipping range, respectively. 
            To emulate communication unreliability, the remote stream undergoes stochastic signal-level dropout (RD): 
            with probability $P_{\text{drop}}=0.1$, its input is replaced by a zero tensor for a duration of at least $T_d \sim \mathcal{N}(2.0,\,0.1)~s$, after which there is a probability $P_{\text{reconnect}}=0.1$ for the camera to reconnect.
            The binary flag $b_t$ indicates this state, enabling the policy to reweight visual cues adaptively.
    
    \subsection{Policy Architecture}
        Both the teacher and student policy networks share the same general architecture consisting of three main modules: a depth encoder, a ROA-based EE~\cite{Fu2022, Agarwal2023} and an actor head, presented in \figref{fig:architecture_overview}.
        While both policies use the same EE and actor head, their main difference lies in the depth encoder:        
        The teacher network uses a scandot-based encoder with access to full terrain geometry, while the student network relies on a depth encoder that uses egocentric and exocentric depth images streams.
        To be able to use both depth streams, the student network encodes both depth inputs using separate convolutional neural network backbones, then combines them with proprioception through a multilayer perceptron (MLP), before going through a gated recurrent unit (GRU) network that preserves temporal consistency. Finally, a last MLP module outputs the depth latent representation, while also estimating the oracle heading~$\boldsymbol{\Delta \psi}$.
        
        In both networks, the depth latent representation is concatenated with the oracle heading $\boldsymbol{\Delta \psi}$ (privileged or estimated), proprioceptive features $\mathbf{o}^{\text{prop}}$, and the extrinsic latents estimated by the EE module.
        The resulting combined vector is then given to the actor module, which generates the corresponding joint position commands.
    
    \subsection{Training Procedure}
        Training proceeds in two stages.
        In Phase~1, the teacher policy is optimized using Proximal Policy Optimization (PPO)~\cite{Schulman2017}.
        The reward is the same as the one defined in \cite{Cheng2024}, which extends \cite{Agarwal2023, Cheng2023}, balancing two tracking and eleven regularization terms, presented in \tabref{tab:reward_terms}.
        
        \begin{table}[htbp!]
            \begin{center}
            \caption{Reward terms and corresponding expressions.} \label{tab:reward_terms}
            \renewcommand{\arraystretch}{1.2}
            \setlength{\tabcolsep}{4pt}
            \begin{tabular}{p{0.3\linewidth} p{0.6\linewidth}}
                \toprule
                    \textbf{Reward term} & \textbf{Expression} \\ 
                \midrule
                    Target vel. tracking &
                    \( 
                     r_{\text{vel\_xy}} = 
                    \min\!\left(\left\langle \mathbf{v}_{\text{xy}}, \frac{\mathbf{g}_{\text{xy}} - \mathbf{p}_{\text{xy}}}
                    {\lVert \mathbf{g}_{\text{xy}}-\mathbf{p}_{\text{xy}}\rVert_2} 
                    \! \right\rangle,\, \text{v}_{\text{cmd}}\right) \) \\
                    
                    Yaw tracking &
                    \( 
                         r_{\psi} = 
                        \exp(-|\Delta \psi_\text{g}|)
                    \) \\
                    
                    Z velocity &
                    \( 
                         r_{\text{vel\_z}} = 
                       \text{v}_z^2
                    \) \\
                    
                    Ang. vel. (roll/pitch) &
                    \( 
                         r_{\omega\text{\_xy}} = 
                        \lVert\boldsymbol{\omega}_{\text{xy}}\rVert_2^2 
                    \) \\
                    
                    Joint error &
                    \( 
                         r_{\Delta q} = 
                       \lVert \mathbf{q} - \mathbf{q_0}\rVert_2^2 
                    \) \\
                    
                    Joint accel. &
                    \( 
                         r_{\ddot{q}} = 
                        \lVert\mathbf{\ddot{q}}\rVert_2^2 
                    \) \\
                    
                    Hip position &
                    \( 
                         r_{\text{hip}} = 
                       \sum_{i\in\mathcal{H}} q_i^2
                    \) \\
                    
                    Collision penalty &
                    \( 
                         r_{\text{collision}} = 
                        \sum_{i \in \mathcal{C} \cup \mathcal{T}} (\lVert \mathbf{f}_i \rVert_2 \ge 0.1N) 
                    \) 
                    \\
                    
                    Action rate &
                    \( 
                         r_a = 
                        \lVert \mathbf{a}_{\text{k}} - \mathbf{a}_{\text{k}-1} \rVert_2
                    \) \\
                    
                    $\Delta$ torques &
                    \( 
                         r_{\Delta \tau} = 
                       \lVert \boldsymbol{\tau_{\text{k}}} - \boldsymbol{\tau_{\text{k}-1}}\rVert_2^2 
                    \) \\
                    
                    Joint torques &
                    \( 
                         r_\tau = 
                        \lVert \boldsymbol{\tau_{\text{k}}} \rVert_2^2
                    \) \\
                    
                    Feet stumble &
                    \( 
                         r_{\text{stumble}} = 
                        \text{any}(\lVert {\mathbf{f}_{i}}_{\text{xy}} \rVert_2 \ge 4 \, {\mathbf{f}_{i}}_{\text{z}}) \biggr\rvert _{i \in \mathcal{F}}
                    \) \\
                    
                    Feet edge &
                    \( 
                         r_{\text{clearance}} = 
                        \sum_{i=0}^{4} C_i M(\mathbf{p}_i) 
                    \) \\
                \bottomrule
            \end{tabular}
            \end{center}
            Where:
            \begin{itemize}
                \item $\mathbf{g}_{\text{xy}}, \, \mathbf{p}_{\text{xy}} \in \mathbb{R}^{2}$: planar (\(X,Y\)) coordinates of the next waypoint and the robot CoM.
                \item $\mathbf{v}_{\text{xy}} \in \mathbb{R}^{2},\, \text{v}_{\text{z}} \in \mathbb{R}$: linear velocity in the horizontal plane and along the vertical axis.
                \item $\text{v}_{\text{cmd}} \in \mathbb{R}$: commanded (target) forward velocity.
                \item $\langle \mathbf{i},\mathbf{j} \rangle$: inner product between vectors \(\mathbf{i}\) and \(\mathbf{j}\). 
                \item $\lVert \mathbf{i}\rVert_2$: Euclidean norm of vector \(\mathbf{i}\).
                \item $\boldsymbol{\omega}_{\text{xy}} \in \mathbb{R}^{2}$: angular velocity around the \(X\) and \(Y\) axes.
                \item $\mathbf{q}, \mathbf{q_0} \in \mathbb{R}^{12}$: current and initial joint configurations.
                \item $\mathbf{\ddot{q}} \in \mathbb{R}^{12}$: joint accelerations.
                \item $\mathcal{H}$, $\mathcal{C}$, $\mathcal{T}$: index sets for hip, calf, and thigh joints.
                \item $\mathbf{f}_i \in \mathbb{R}^{3}$: contact force applied to the \(i^\text{th}\) rigid body, with ${\mathbf{f}_{i}}_{\text{xy}}\in\mathbb{R}^{2}, \, {\mathbf{f}_{i}}_{\text{z}}\in\mathbb{R}$ are respectively its horizontal and vertical components.
                \item $\mathbf{a}_{\text{k}}, \mathbf{a}_{\text{k}-1} \in \mathbb{R}^{12}$: actions vectors at the current and previous control steps.
                \item $\boldsymbol{\tau}_{\text{k}}, \boldsymbol{\tau}_{\text{k}-1} \in \mathbb{R}^{12}$: joints torques at the current and previous control steps.
                \item $\mathcal{F}$: index sets for the feet rigid bodies.
                \item $C_i$: binary contact indicator of the \(i^\text{th}\) foot (1 if in ground contact, 0 otherwise).
                \item $M(\mathbf{X})$: Boolean function returning 1 if point \(\mathbf{X}\) is within 5~cm of an edge, and 0 otherwise.
                \item $\mathbf{p}_i$: 3D position of the \(i^\text{th}\) foot.
            \end{itemize}
        \end{table}

        The total reward is then defined as a proportional sum:
            \begin{equation} \label{eq:total_reward}
                \begin{split}
                    r_{tot} =  
                        & 1.5 \, r_{\text{vel\_xy}} + 0.5 \,  r_{\psi} - r_{\text{vel\_z}} - 0.05 \, r_{\omega\text{\_xy}} - 0.04 \, r_{\Delta q}\\
                        &   - \, r_{\ddot{q}}  - 0.5 \,  r_{\text{hip}} - 10 \, r_{\text{collision}} - 0.1 \, r_{a}  - 10^{-7} \, r_{\Delta \tau} \\
                        & - 10^{-5} \, r_\tau
                        -  r_{\text{stumble}} -  r_{\text{clearance}} 
                \end{split}
            \end{equation} 
        The reward sum is clipped at 0 for negative values to avoid early termination.
        
        In Phase~2, the student policy is distilled via DAGGER~\cite{Ross2011}, minimizing the following distillation loss:  
        \begin{equation}
            \mathcal{L}_{\text{distill}} = \lVert \mathbf{a}^{\text{S}} - \mathbf{a}^{\text{T}} \rVert_2 + \lVert {\boldsymbol{\Delta \psi}}^{\text{S}} - {\boldsymbol{\Delta \psi}}^{\text{T}} \rVert_2
        \end{equation}
        where ${\boldsymbol{\Delta \psi}}^{\text{S}}, \, \boldsymbol{\Delta \psi}^{\text{T}}\in\mathbb{R}^{2}$ 
        denote respectively the student and teacher estimated/privileged oracle heading and $\mathbf{a}^{\text{S}},\,\mathbf{a}^{\text{T}}\in\mathbb{R}^{12}$ denote respectively the student and teacher actions.
        All perturbations, i.e. orientation, positional and dropout, remain active throughout.

%% file: Sections/results.tex
\section{Results} \label{sec:results}
    
    We evaluate the proposed multi-viewpoints locomotion policy using two experimental sets, each with the goal of validating a different criteria:
    (1)~the benefit of incorporating exocentric vision, and 
    (2)~the robustness of the combined vision controller when the remote camera sustains spatial displacement. 
    All policies are trained and tested under the same domain-randomized conditions described in \secref{subsec:training_environment}. 
    Results are averaged over 256 parallel runs across 25 procedurally generated terrains of 4 different types (hurdles, steps, gaps and parkours\footnote{A combination of hurdles, steps, gaps and slopes.}), for 1500 iterations. The code used to train the models and obtain the results is openly available at \url{https://anonymous.4open.science/r/multiview-parkour-6FB8}.
    
    \subsection{Effect of Vision Configuration}        
        The first set of experiments investigates how different visual inputs affect locomotion performance.
        We compare four policies:
        (1)~an onboard-only baseline policy equivalent to~\cite{Cheng2024},
        (2)~a remote-only policy,
        (3)~a combined vision policy trained without random remote-camera dropouts (RD), and
        (4)~a combined vision policy trained with random RD. 

\begin{figure}[htbp!]
            \centering
            \includegraphics[width=0.45\linewidth]{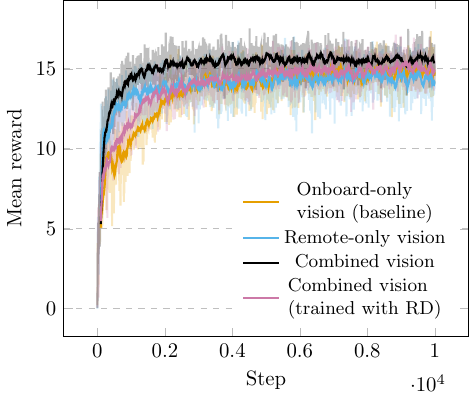}
            \caption{Mean training reward across all models. A moving average with a window size of 100 is applied for clarity.}
            \label{fig:student_training}
        \end{figure}

        Training for all models is done using the orientation and dropout perturbations described in \secref{sec:method}. After \num{10000} iterations, the policies converge toward an optimal solution, even under sensory degradation, as can be seen in \figref{fig:student_training}.

        For evaluation, the policies using remote vision (\textit{Remote-only vision} policy and \textit{Combined vision} policies) are evaluated with and without RD.
        Remote signal loss is performed similarly as during training (\secref{subsec:observation_and_input_representation}): with probability $P_{\text{drop}}=0.1$, the remote camera input is replaced by a zero tensor for a duration of at least $T_d \sim \mathcal{N}(2.0,\,0.1)~s$, after which there is a probability $P_{\text{reconnect}}=0.1$ for the camera to reconnect.
        Finally, a \textit{full disconnect} scenario is also evaluated for the two \textit{Combined vision} policies:
        the remote-view camera is disconnected at the beginning of the experiment (replaced by a zero tensor), and never reconnected, i.e. probability $P_{\text{drop}}=1.0$, and probability $P_{\text{reconnect}}=0.0$. The goal of this last scenario is to assess the performance of the policies using combined vision in a situation similar to the baseline. 
        
        To evaluate the performance of the different policies, we use five metrics: 
        (1)~the \textit{average episode reward}~(AER) is the computed average value of the reward defined in \neqref{eq:total_reward};
        (2)~the \textit{traversal success rate}~(TSR) represents the proportion of completed courses without failure over the 1500 iterations;
        (3)~the \textit{mean x-displacement}~(MXD), normalized to $[0, 1]$, represents forward progression by computing the average proportion of waypoints reached,
        as defined in \cite{Cheng2024};
        (4)~the \textit{mean edge violation}~(MEV), also defined in \cite{Cheng2024}, represents the policy's foothold precision by averaging the number of foot contacts occurring on the edges of the terrain, where lower values denote safer steps; and
        (5)~the \textit{average dropout time}~(ADT) represents the average time during which the remote camera was unavailable for each disconnection session.
        While this last metric cannot be used for direct performance comparison, it is useful to make sure the dropout conditions are similar between each policy evaluation.
        The results can be found in \tabref{tab:vision_comparison}.

        \begin{table*}[htbp!]
            \caption{Comparison of vision configurations under varying signal conditions.} \label{tab:vision_comparison} 
            \begin{minipage}{\textwidth}
                \begin{center}
                    \input{tables/Table_comparison}\par
                \end{center}
                
                Offline evaluation across models. RD (val.): remote-camera dropout during validation. 
                AER = Average Episode Reward (↑ higher is better); TSR = Traversal Success Rate (↑); 
                MXD = Mean X-Displacement, normalized (↑); MEV = Mean Edge Violation (↓); 
                ADT = Average Disconnection Time. $^{*}$ Full-disconnect scenario (remote camera set to zero for the entire episode).
    
            \end{minipage}
        \end{table*}
        
        Across all metrics, the \textit{combined vision model trained with RD} achieves the best balance between performance, stability and adaptability. 
        It matches the baseline's TSR and MXD when the remote view is fully disconnected, demonstrating learned fallback behavior, while outperforming it when the remote camera is available. 
        
        In contrast, the \textit{remote-only variant} and the \textit{combined vision model trained without RD} both perform well when vision is uninterrupted, but exhibits degradation during unexpected remote-camera dropouts. 
        This behavior was expected from the remote-only variant, as it can only rely on proprioception during RD, thus being unable to anticipate and adapt to its environment, only react to it.
        
        In the case of the model trained without RD, these results suggest that the model relies a lot on exocentric vision, hinting to the importance of such viewpoints in environmental understanding.
        They also reinforce the need to include signal-loss events in training to increase robustness to life-like scenarios, which could include such data-stream disconnections.
        
        Overall, these findings confirm that multi-view fusion enhances situational awareness and locomotion precision while maintaining resilience under partial sensory failure, when those are taken into account during training.
        
    \subsection{Qualitative Observations}
    
\begin{figure*}[tb!]
            \centering
            \begin{subfigure}[t]{\linewidth}
                \centering
                \includegraphics[width=0.98\linewidth]{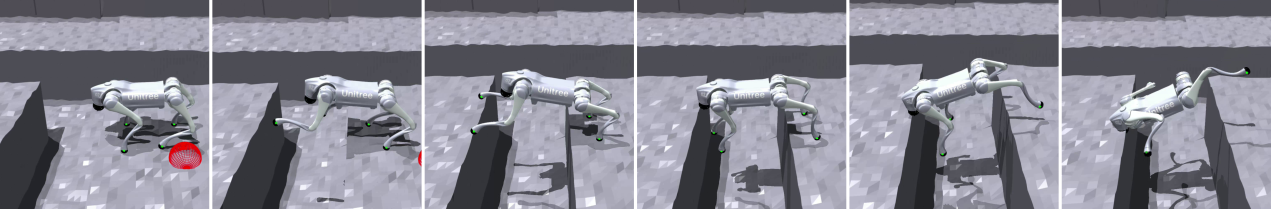}
                \caption{Baseline model crossing a gap.} \label{subfig:qualitative_baseline_vs_combined:baseline}
            \end{subfigure}
            \begin{subfigure}[t]{\linewidth}
                \centering
                \includegraphics[width=0.98\linewidth]{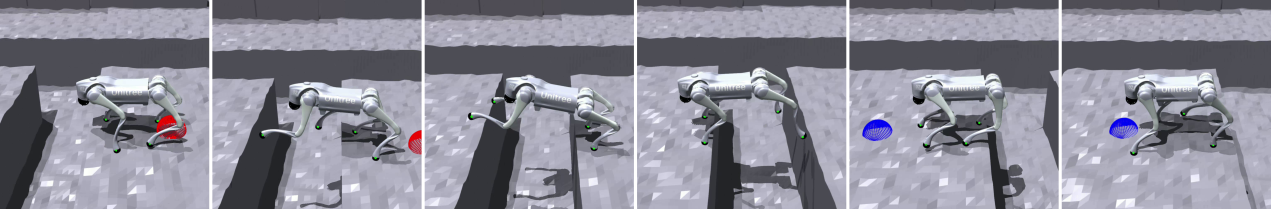}
                \caption{Combined vision model crossing a gap.} \label{subfig:qualitative_baseline_vs_combined:combined}
            \end{subfigure}
            \begin{subfigure}[t]{\linewidth}
                \centering
                \includegraphics[width=0.98\linewidth]{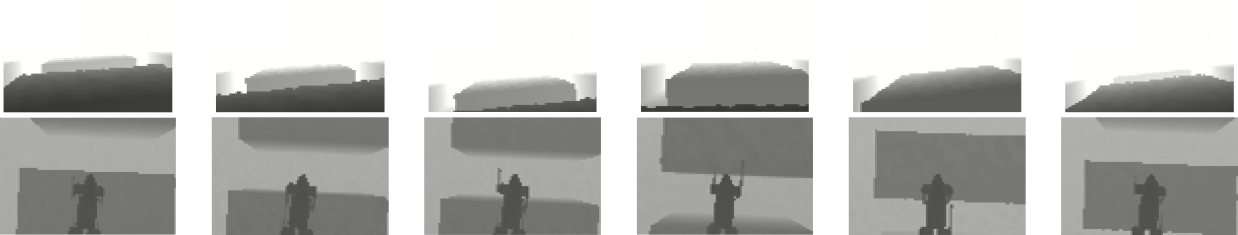}
                \caption{Egocentric (top) and exocentric (bottom) depth data, as seen by the combined vision model in (b). Data streams are delayed by 20~ms.} \label{subfig:qualitative_baseline_vs_combined:combined_depth}
            \end{subfigure}
            \caption{Comparison of the baseline model (a) and the combined vision model trained without RD (b) when executing a locomotion sequence to jump over a gap, using the perception data shown in (c). The baseline model (a) underestimates the gap distance, resulting in an insufficient jump. In contrast, the combined-vision model (b) accurately estimates the required distance and successfully performs the jump.}
            \label{fig:qualitative_baseline_vs_combined}
        \end{figure*}

        To complement the quantitative results, \figref{fig:qualitative_baseline_vs_combined} compares locomotion sequences produced by the \textit{baseline} egocentric policy and the \textit{combined vision policy trained without RD} during a gap-crossing task.
        The robot relying solely on egocentric vision (\figref{subfig:qualitative_baseline_vs_combined:baseline}) misjudges the gap distance, initiating an early and insufficiently long jump that leads to failure.
        In contrast, the combined vision policy (\figref{subfig:qualitative_baseline_vs_combined:combined}) accurately estimates the gap size using both ego- and exocentric viewpoints, executing a stable and well-timed leap.

\begin{figure*}[tb!]
            \centering
            \begin{subfigure}[t]{\linewidth}
                \centering
                \includegraphics[width=0.98\linewidth]{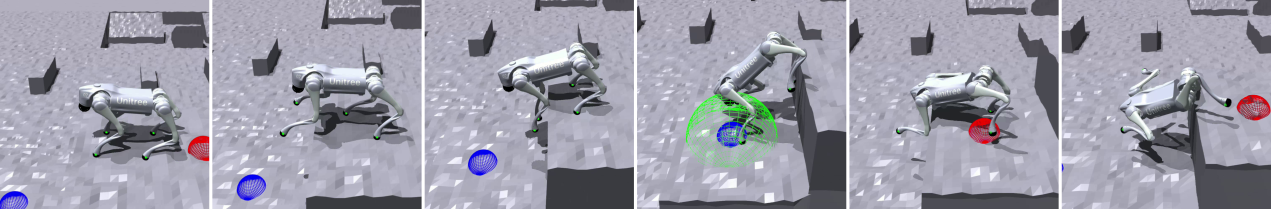}
                \caption{Combined vision model trained without RD going down steps while suffering RD.} \label{subfig:qualitative_combined_vs_combined:without_RD}
            \end{subfigure}
            \begin{subfigure}[t]{\linewidth}
                \centering
                \includegraphics[width=0.98\linewidth]{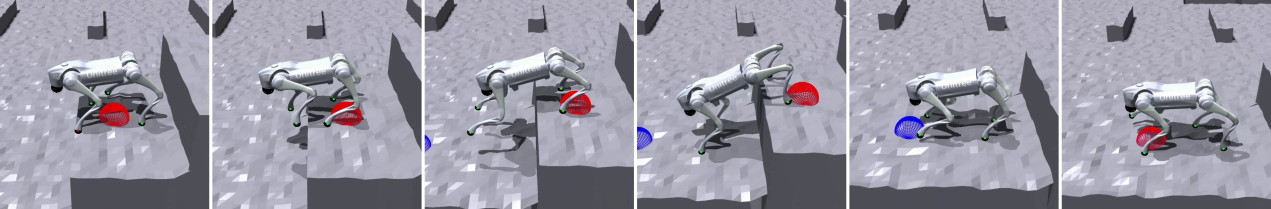}
                \caption{Combined vision model trained with RD going down steps while suffering RD.} \label{subfig:qualitative_combined_vs_combined:with_RD}
            \end{subfigure}
            \caption{Comparison between the combined-vision model trained without (a) and with RD (b) when executing a locomotion sequence to descend a step while experiencing RD. In the model trained without RD (a), the robot over-relies on the exocentric view, leading to catastrophic failure when RD occurs. In contrast, the model trained with RD (b) learns to integrate both viewpoints robustly, enabling safe and stable descent even under RD conditions.} \label{fig:qualitative_combined_vs_combined}
        \end{figure*}

        To assess the impact of RD, \figref{fig:qualitative_combined_vs_combined} illustrates how combined vision policies trained with and without RD behave when experiencing temporary camera disconnections while descending steps.
        When the remote camera signal is lost, the policy trained without RD (\figref{subfig:qualitative_combined_vs_combined:without_RD}) exhibits erratic motion, suggesting overreliance on exocentric information and limited capacity to compensate using only egocentric information.
        On the other hand, the RD-trained policy (\figref{subfig:qualitative_combined_vs_combined:with_RD}) maintains stable gait and balance, seamlessly shifting to egocentric view.
        This adaptive behavior highlights the importance of including RD during training: it teaches the policy to dynamically reweight sensory inputs according to their availability, improving resilience in the face of intermittent input failure.

    \subsection{Effect of Remote-View Displacement}
        The second set of experiments assesses the robustness of the policy under increasing spatial misalignment between the robot and the remote camera, following the spherical perturbation model described in \secref{subsec:training_environment}. 
        To do so, we train six combined vision policies, each using random remote camera displacements with a distinct maximum displacement radius \( R_s^{t} \in \{0.0, 0.1, 0.2, 0.3, 0.4, 0.5\} \).
        Each policy is then evaluated under six different maximum displacement radii \( R_s^{e} \in \{0.0, 0.1, 0.2, 0.3, 0.4, 0.5\} \), resulting in a comprehensive cross-evaluation of training and testing misalignment conditions. 

\begin{figure}[htbp]
            \centering
            \begin{subfigure}[t]{0.49\textwidth}
                \includegraphics[width=\linewidth]{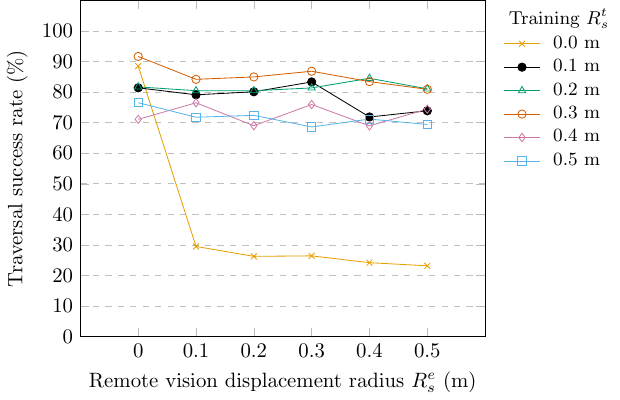}
                \caption{Traversal success rate of 6 models.}
                \label{fig:success_rate_vs_rs}
            \end{subfigure}
            \begin{subfigure}[t]{0.49\textwidth}
                \includegraphics[width=\linewidth]{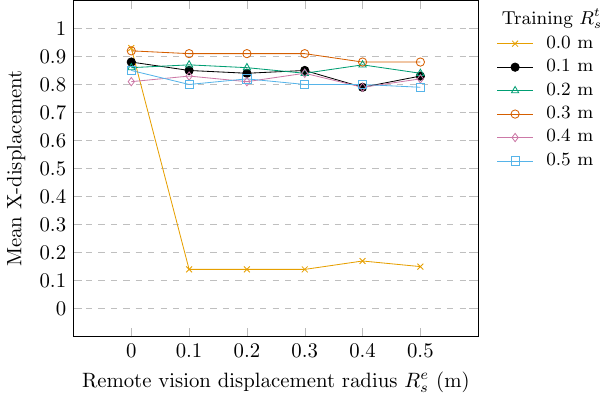}
                \caption{Mean normalized x-displacement of 6 models. The average standard deviation is around 0.27.}
                \label{fig:xdisp_vs_rs}
            \end{subfigure}
            \caption{Traversal success rate (a) and mean normalized X-displacement (b) of 6 models, trained with different remote camera displacement radius $R_s^t$, as a function of remote camera displacement radius $R_s^e$}
        \end{figure}

        \figref{fig:success_rate_vs_rs} shows the traversal success rate as a function of displacement radius $R_s^e$, while \figref{fig:xdisp_vs_rs} presents the corresponding normalized mean x-displacement. 
        
        The results show that the model trained without any positional variation ($R_s^t = 0.0~\mathrm{m}$) suffers a degradation when tested under misaligned viewpoints, with TSR dropping from almost $90\%$ at perfect alignment to below $30\%$ with only an offset of $R_s^e = 0.1~\mathrm{m}$. 
        This behavior suggests an overfit to static camera alignment and an inability to compensate for viewpoint inconsistencies.

        On the other hand, training with larger perturbations ($R_s^t \ge 0.4~\mathrm{m}$) provides some regularization benefits but slightly reduces maximum performance and increases variability. 
        The best overall trade-off occurs around $R_s^t = 0.3~\mathrm{m}$, where TSR remains above $80\%$ and MXD stays near its maximum across all test conditions. 
        This suggests that moderate viewpoint randomization effectively encourages the policy to develop robustness to positional disturbances without affecting its spatial representation. 
        These results indicate that, when taken into account during training, modest remote-view instability can be tolerated without significant performance loss, an important ability for deployment on aerial-ground cooperative systems where remote camera position stability cannot be guaranteed.

%% file: tables/Table_comparison.tex
\newcommand{\NA}{\textit{N/A}}
\newcommand{\best}[1]{\textbf{#1}}
\newcommand{\cmark}{\ding{51}}           
\newcommand{\numpm}[2]{\num{#1}\,$\pm$\,\num{#2}}
\centering
\setlength{\tabcolsep}{6pt}
\renewcommand{\arraystretch}{1.05}
\begingroup
\begin{tabular}{l c c c c c c}
\toprule
\makecell{\textbf{Model}} &
\makecell{\textbf{RD}\\\textbf{(val.)}} &
\makecell{\textbf{AER}\\(\(\uparrow\))} &
\makecell{\textbf{TSR}\\(\% \(\uparrow\))} &
\makecell{\textbf{MXD}\\(\(\uparrow\))} &
\makecell{\textbf{MEV}\\(\(\downarrow\))} &
\makecell{\textbf{ADT}\\(s)} \\
\midrule

\makecell{Baseline \cite{Cheng2024}\\(Onboard-only vision)} &  &
  \numpm{13.66}{6.24} & \num{61.02}\% & \numpm{0.73}{0.31} & \numpm{0.03}{0.16} & \NA \\

\midrule

\multirow{2}{*}{\makecell{Remote-only vision}} &  &
  \numpm{17.30}{5.90} & \best{\num{91.84}}\% & \numpm{0.90}{0.23} & \numpm{0.02}{0.15} & \NA \\
& \cmark &
  \numpm{10.18}{5.29} & \num{41.60}\% & \numpm{0.55}{0.28} & \numpm{0.08}{0.30} & \numpm{1.61}{1.12} \\

\midrule
\multirow{3}{*}{\makecell{Combined vision}} &  &
  \best{\numpm{18.97}{5.60}} & \num{90.24}\% & \best{\numpm{0.93}{0.19}} & \best{\numpm{0.01}{0.08}} & \NA \\
& \cmark &
  \numpm{8.06}{3.63} & \num{28.23}\% & \numpm{0.43}{0.25} & \numpm{0.05}{0.21} & {\numpm{1.49}{0.89}} \\
& \cmark$^{*}$ &
  \numpm{3.52}{1.35} & \num{23.19}\% & \numpm{0.21}{0.12} & \numpm{0.09}{0.32} & \NA \\

\hdashline

\multirow{3}{*}{\makecell{Combined vision\\(RD during learning)}} &  &
  \numpm{18.72}{6.06} & \num{87.64}\% & \best{\numpm{0.93}{0.19}} & \best{\numpm{0.01}{0.11}} & \NA \\
& \cmark &
  \numpm{17.74}{6.24} & \num{78.77}\% & \numpm{0.89}{0.23} & \numpm{0.03}{0.21} & \numpm{1.58}{1.17} \\
& \cmark$^{*}$ &
  \numpm{14.35}{6.58} & \num{63.86}\% & \numpm{0.74}{0.32} & \numpm{0.04}{0.22} & \NA \\
\bottomrule
\end{tabular}
\endgroup

%% file: Sections/conclusion.tex
\section{Conclusion} \label{sec:conclusion}
    This work introduced a multi-view visual locomotion framework that combines egocentric and exocentric depth perception to enhance quadrupedal traversal across discontinuous and uncertain terrains. 
    Building upon prior visuomotor control methods, the proposed approach integrates complementary visual perspectives to enable more anticipatory and stable motion strategies. 
    Extensive simulation experiments demonstrate that integrating exocentric perception significantly enhances locomotion performance compared to purely egocentric systems. 
    Training under intermittent disconnections produces policies that are both adaptive and fault-tolerant, effectively leveraging remote observations when available while relying on onboard vision and proprioception during outages. 
    The spherical perturbation analysis further presents the operational bounds for camera alignment, highlighting the practical feasibility of aerial-ground cooperative deployments. 
    These findings reinforce the importance of incorporating multiple viewpoints within the locomotion pipeline and highlight the role of extensive domain randomization in producing adaptive and resilient policies capable of bridging the sim-to-real gap by leveraging spatially diverse yet intermittently unreliable sensory inputs.
    
    Future work will focus on transferring the proposed system to real hardware, investigating temporal synchronization across asynchronous viewpoints, and extending the framework to heterogeneous multi-robot teams. 
    By enabling more adaptive, perceptually rich locomotion, this study moves one step closer to autonomous legged systems capable of operating with greater agility and developing a better spatial awareness in natural environments.